\documentclass[runningheads]{llncs}
\usepackage{times}
\usepackage{epsfig}
\usepackage{graphicx}
\usepackage{amsmath}
\usepackage{amssymb}
\usepackage[lined,algonl,ruled]{algorithm2e}
\usepackage{paralist}
\usepackage{color}
\usepackage{hyperref}
\usepackage{flushend}

\begin{document}

\title{Forming Local Intersections of Projections for Classifying and Searching Histopathology Images}

\author{A. Sriram\inst{1}\and S. Kalra\inst{1}\and  M. Babaie\inst{1}\and B. Kieffer\inst{1}\and W. Al Drobi\inst{1}\and S. Rahnamayan\inst{2}\and H. Kashani\inst{1}\and H.R. Tizhoosh\inst{1}}
\authorrunning{To appear in International Conference on AI in Medicine (AIME 2020)}

\institute{KIMIA Lab, University of Waterloo, Canada \and
 Elect., Comp. and Software Eng., Ontario Tech University, Canada}

\maketitle

\begin{abstract}
In this paper, we propose a novel image descriptor called ``Forming Local Intersections of Projections'' (FLIP) and its multi-resolutional version (mFLIP) for representing histopathology images. The descriptor is based on the Radon transform wherein we apply parallel projections in small local neighborhoods of gray-level images. Using equidistant projection directions in each window, we extract unique and invariant characteristics of the neighborhood by taking the intersection of adjacent projections. Thereafter, we construct a histogram for each image, which we call the FLIP histogram. Various resolutions provide different FLIP histograms which are then concatenated to form the mFLIP descriptor. Our experiments included training common networks from scratch and fine-tuning pre-trained networks to benchmark our proposed descriptor. Experiments are conducted on the publicly available dataset KIMIA Path24 and KIMIA Path960. For both of these datasets, FLIP and mFLIP descriptors show promising results in all experiments.Using KIMIA Path24 data, FLIP outperformed non-fine-tuned Inception-v3 and fine-tuned VGG16 and mFLIP outperformed fine-tuned Inception-v3 in feature extracting.
\end{abstract}

\keywords{
Radon Projections Histopathology \and
Image Search \and
Feature Extraction \and
Image Descriptors.}

\section{Introduction}
 Histopathology, is primarily concerned with the manifestations of a diseased human tissue \cite{lever1949histopathology}. A traditional diagnosis is based on examination of the tissue of concern mounted on a glass slide under various magnifications of a microscope. \cite{kayser1984pattern}. More recently, digital pathology has connected the computer vision field to the diagnostic pathology by scanning the glass slide and creating a whole slide image (WSI). This allows easy storage, more flexibility in sharing information and   eliminates the risk of losing specimens \cite{farahani2015whole}. In digital pathology the images are extremely large and it takes an extremely long time to compute images. Hence, there is a need to develop a powerful image descriptor that can extract unique and invariant features from these large images to enable algorithms to retrieve and classify salient patterns and morphologies. In essence, the descriptor should suffice as an image representative such that one should be able to index the large scans with a limited number of descriptors. 

In this work, we propose a novel descriptor called ``Forming Local Intersections of Projections'' (FLIP) which applies the Radon transform to small spatial windows (to extract the features of histopathology images). The FLIP descriptor enables fast image search while minimizing any extra storage requirement by storing the representation of an image as a compact histogram. 
\section{Related Works}
\label{sec:lit_rev}
Image descriptors quantify image characteristics such as shape, color, texture, edges, and corners. Local Binary Patterns (LBPs) \cite{ojala2002multiresolution} are a good example for image descriptor to classify texture with rotation invariance \cite{pietikainen2011local}. Designed as a particular case of texture spectrum model \cite{ojala1996comparative}, LBPs are powerful image descriptors that have certainly set relatively high accuracy standards in the medical domain, including  digital pathology scans \cite{babaie2017classification}. 

Deep neural networks have  been widely utilized to generate global image descriptors.  These networks consist of functions in each layer to generate local features at different resolutions describing a particular image region. These local features may then be aggregated,to provide a global descriptor that is the entire image. Similar to LBP, deep descriptors have reported many promising results, specifically in the histopathology domain \cite{xu2014deep,babaie2017classification}. 

More recently, several approaches have been put forward to develop projection-based descriptors \cite{elppaper,babaie2017local}.The Radon transform is a well-established approach \cite{sanz2013radon}. A novel Radon barcode for medical image retrieval system was proposed in 2015 \cite{tizhoosh2015barcode}. The Radon barcode is a binary vector generated from global projections with selected projection angles and projection binarization operation that can tag a medical image or its regions of interest. Using Radon barcodes, large image archives can be efficiently searched to find matches via Hamming distance,however, the performance of global projections is rather limited. More recently, local Radon projections and support vector machines (SVM) have been combined for medical image retrieval  \cite{elppaper}. 


Tizhoosh et. al \cite{tizhoosh2016barcodes} have introduced Autoencoded Radon Barcode (ARBC) that used mini-batch stochastic gradient descent and binarizing the outputs from each hidden layer during training to produce a barcode per-layer. The ARBC was observed to achieve an Image Retrieval in Medical Application (IRMA) error of 392.09. More recently, Tizhoosh et al. \cite{tizhoosh2016minmax} proposed MinMax Radon barcodes which were observed to retrieve images 15\% faster compared to the ``local thresholding'' method. 

A similar type of approach was proposed by Xiaoshuang et al. \cite{shi2017cell} which presented a cell-based framework for pathology images wherein they encode each cell into a set of binary codes using a hashing model \cite{shi2017supervised}. The binary codes are then converted into a 1-dimensional histogram vector (which is the feature vector) used for learning using an SVM for image classification. 
In this paper, we attempt to design and test a `` local'' projection-based descriptor that should deliver good results for histopathology images. 

\section{Methods}
\label{sec: FLIP}




The Radon transform provides scene/object projections (profiles) in different directions. The set of all projections can yield a reconstruction of the scene/objects when performing an inverse Radon transform (i.e., filtered backprojection). 

Using the Dirac delta function $\delta(\cdot)$, the Radon transform of a two-dimensional image $f(x, y)$ can be defined as its line integral along a straight line inclined at an angle $\theta$ and at a distance $\rho$ from the origin: 

\begin{equation}
R(\rho, \theta) = \int_{-\infty}^{\infty} \int_{-\infty}^{\infty} f(x, y)\delta(x cos\theta + y sin\theta - \rho) dx dy 
\end{equation}

Here, $-\infty < \rho < \infty, 0 \leq \theta < \pi$, the Radon transform accentuate straight-line features from an image by integrating the image intensity over the straight lines to a single point \cite{rey1990application}. Given the scan $\mathbf{S}$, we are interested in describing the  (grayscale) image  $\mathbf{I}(\subset \mathbf{S})$ via a short descriptor, or histogram, $\mathbf{h}$ using Radon projections $R(\rho,\theta)$ to transform the intensities $f(x,y)$ of $\mathbf{I}$. We can process all local neighbourhoods $\mathbf{W}_{ij}\subset \mathbf{I}$. For each neighbourhood $\mathbf{W}_{ij}$, we capture $n_P$ projections with $0<n_P\ll 180$: $\mathbf{p}^1_{ij},\mathbf{p}^2_{ij},\dots,\mathbf{p}^{n_P}_{ij}$. One may find individual projections from different (and dissimilar) images to be quite similar. Hence, we take the ``\emph{intersection of adjacent projections}'' to quantify the spatial correlations of a given neighbourhood pattern. The intersection of projections can be thought of as an approximation of the logical ``AND'' providing a unique characteristic of local patterns. Therefore, we receive $n_P$ intersection vectors $\mathbf{V}_{k,m}$ as 
\begin{equation}
\mathbf{V}_{k,m}=\min\left(\mathbf{p}^k_{ij},\mathbf{p}^{(k+1)\%n_P}_{ij}\right), 
\end{equation}
with $k=1,2,\dots,n_P$ and $m=1,2,\dots, n_W$ where $n_W$ is the total number of local windows of the image $\mathbf{I}$. Hence, we will have $n_P\times n_W$ intersections of local projections. The projections have different values which are also subject to intensity fluctuations. Hence we re-scale all projection values to be:
\begin{equation}
\mathbf{\bar{V}}_{k,m}=\left\lceil {L\times\frac{\mathbf{V}_{k,m}-p_{\min}}{p_{\max}-p_{\min}}}\right\rceil.
\end{equation}

\begin{figure}[htb]
	\centering
	\includegraphics[scale=0.22]{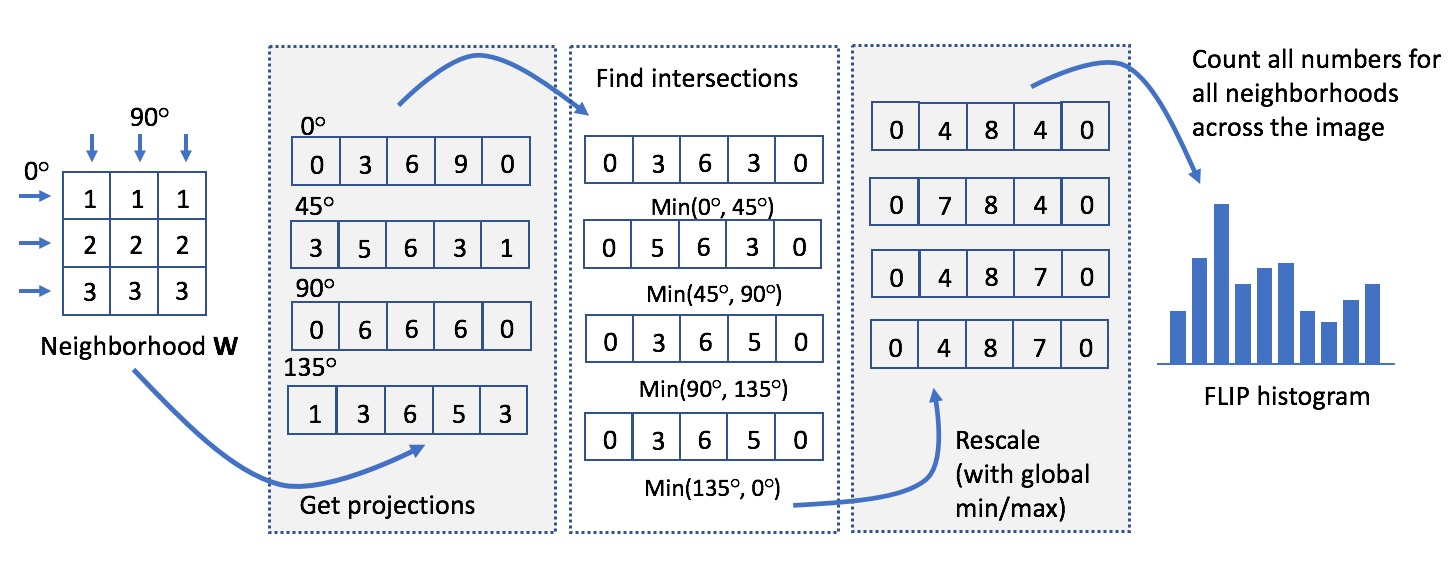} 
	\caption{A simplified overview of the FLIP histogram extraction. A small number of pairwise orthogonal projections, here $n=4$, is computed for each neighbourhood, from which the intersection of all adjacent projections is computed. After re-scaling all intersections in the image based on global min/max projection values, the FLIP histogram can be assembled by counting the rescaled intersections. }
	\label{fig:FLIP_breakdown}
\end{figure}

Now, we can count the values $\mathbf{\bar{V}}\in{1,2,\dots,L},\forall m=1,2,\dots,n_P\times n_W$ to obtain $\mathbf{h}$, wherein $\mathbf{L}$ is the default histogram length of $128$. Algorithm \ref{algo: FLIP} provides the pseudo-code for calculating the FLIP descriptor. Figure \ref{fig:FLIP_breakdown} provides a simplified overview of extracting a FLIP histogram for sample images.  

\begin{algorithm}[t]
	\small
	\SetKwInOut{Input}{Input}
	\SetKwInOut{Output}{Output}
	\Input{An image $\mathbf{I}$ as part of a whole scan $\mathbf{S}$: $\mathbf{I}\subset \mathbf{S}$ }
	\Output{The FLIP histogram $\mathbf{h}$}
	Set neighbourhood size and overlap\;
	$L \gets 128 $ (histogram default length)\;
	$\mathbf{h} \gets \emptyset$\;
	$\mathbf{F} \gets \emptyset$\;
	$\mathbf{I}_g \gets$ \textrm{Convert image $\mathbf{I}$ to gray-scale}\;
	\BlankLine
	\ForEach{window $\mathbf{W}_i$ in image $\mathbf{I}_g$}{
		$R_{(0,45,90,135)} \gets \textrm{RadonTransform}(\mathbf{W}_i)$\;
		$R_{\min_1} \gets$ \textrm{$\min(R_{0}, R_{45})$}\;
		$R_{\min_2} \gets$ \textrm{$\min(R_{45}, R_{90})$}\;
		$R_{\min_3} \gets$ \textrm{$\min(R_{90}, R_{135})$}\;
		$R_{\min_4} \gets$ \textrm{$\min(R_{135}, R_{0})$}\;
		$R_{\min} \gets$ \textrm{concatenate} $(R_{\min_1}, R_{\min_2}, R_{\min_3}, R_{\min_4})$\;
		$\mathbf{F} \gets \textrm{AppendRow}(R_{\min})$\;    
	}
	$f_{\min}, f_{\max} \gets \textrm{FindMinMax}(\mathbf{F})$\;
	$\mathbf{F} \gets \textrm{reScale}(\mathbf{F},f_{\min}, f_{\max}, L)$\;
	$\mathbf{F} \gets \mathbf{F [1:128]} \textrm{ (127 length histogram)}$\;
	\For{$i=1$ to $\mathbf{F}_\textrm{rows}$}{
		\For{$j=1$ to $\mathbf{F}_\textrm{cols}$}{
			$\mathbf{h}(\mathbf{F}(i,j)) \gets \mathbf{h}(\mathbf{F}(i,j)) + 1$\; 
		}
	}
	Return $\mathbf{h}$\;
	\caption{The FLIP algorithm}
	\label{algo: FLIP}
\end{algorithm}

 The intuition behind multi-resolutional representation is to capture structural changes that one observes in real-world objects \cite{chan2007multi}. Specifically, multiple scales for the same image capture the variation in visual appearances - providing a different representation of the same image for every scale. Furthermore, pathologists examine tissue samples at different magnifications to have a comprehensive perception of the specimen \cite{pantanowitz2010digital}. A multi-resolution FLIP is built using different image resolutions (inclusive of original and resized resolutions). These resolutions include: 
\begin{inparaenum}[(i)]
\item original resolution,
\item 0.75$\times$ the original resolution,
\item 0.5$\times$ the original resolution, and 
\item 0.25$\times$ the original resolution.
\end{inparaenum}
After obtaining a FLIP histogram for each of the resolutions, we concatenate them in descending order of resolution to form a final histogram, namely the mFLIP descriptor. 
 


\textbf{Indexing and Testing --}
For all training samples, the scan is first divided by a regular grid of proper size whereas each grid cell can be cropped into a new training image/patch. The mFLIP histogram of each image is calculated and saved in a database. Subsequently, all of the images are classified using an SVM algorithm. In the testing phase (which emulates the proposed system in action), every query scan can be processed in two possible ways:
\begin{inparaenum}[(i)]
\item a small number of locations within the query scan is selected to extract some patches, or
\item one patch within the query scan is selected either manually or automatically.
\end{inparaenum}
In both cases the main task of the image search is to find the best match for a patch.  The matching algorithm can be implemented in two ways: 
\begin{inparaenum}[(i)]
\item using a proper distance measure, we quantify the (dis)similarity between the mFLIP descriptor of the query patch and the mFLIP descriptor of every image in the database, or
\item we use the trained SVM to assign a class to the query patch.
\end{inparaenum}
For the distance-based image search, several strategies were used including: $\chi^2$, histogram intersection, Pearson coefficient, cosine similarity, and $L_1$ and $L_2$ metrics. As for classification, a generalized histogram intersection kernel SVM is adopted. In practice, pathologists prefer to inspect more than one retrieved case. Hence, we retrieve the top 3 images (patches) for a query scan (for visual inspection) and examine which one of the three images is the actual match for the query image to calculate the accuracy (only the first match is considered for accuracy calculation). 
\section{Experiments}
\label{sec:experiments}

We used two publicly available pathology datasets to validate the FLIP and mFLIP algorithms, namely: (i) KIMIA Path960, and (ii) KIMIA Path24. All images in these experiments are converted to grayscale.

\textbf{KIMIA Path960 - } Introduced by Kumar et al. \cite{kumar2017comparative}. KIMIA Path960 is a publicly available pathology dataset, comprised of 960 images of size 308$\times$168 from 20 different classes (i.e tissue types). Since this dataset is relatively small, we used leave-one-out approach to validate our proposed algorithms. 

\textbf{KIMIA Path24 - } Introduced by Babaie et al. \cite{babaie2017classification}, is a digital pathology public dataset, published in 2017, that comprises of 24 scans depicting different tissue patterns and body parts. The scans have been converted to gray-scale using the Python library Scikit Learn. The dataset consists of 1,325 test patches of size 1000$\times$1000 (0.5mm$\times$0.5mm) for which the labels correspond to the scan number. The number of training patches can range from 27,000 to 50,000 patches, depending on the percentage of overlap selected by the algorithm designer. In our experiments, we have received 27,055 patches with \%0 overlap.

For the proposed algorithms (FLIP and mFLIP), both image search and classification strategies are implemented to support the algorithms performance. For the image search, we compare the FLIP or mFLIP histogram using either chi-square or histogram intersection algorithm to retrieve a patch that is best-matched with the query image. We then use the label for the best-matched patch to determine the accuracy. The retrieved patch does not necessarily reside within the same WSI. Hence, in the KIMIA Path24 dataset, there are two accuracy measures – patch-to-scan accuracy ($\eta_{p}$) and whole-scan accuracy ($\eta_{W}$). The total accuracy ($\eta_{total}$) is a multiplication of both these accuracies. As for the classification, we train an SVM classifier for all the training patches. We down-sampled each image to 250$\times$250 which resulted in 2$\sim$3\% loss in accuracy, regardless of the histogram length, when compared to the accuracy results for the 1000$\times$1000 images. Hence, we only report results for the gray-scaled 1000$\times$1000 images as they yield better results.

\textbf{Accuracy Measurement --} For the KIMIA Path24, a total of $n_{tot} = 1,325$ test patches $P^j_s$ are obtained which belong to either one of the 24 classes available $\Gamma_s = \{P^i_s | s \in S, i = 1, 2, \dots , n_{\Gamma_s}\}$ with $s=0, 1, 2, \dots , 23$ \cite{babaie2017classification}. In order to compare our method against other works, the accuracy calculation outlined in \cite{babaie2017classification} is adopted. Hence, for a retrieved image $R$ for any experiment, the patch-to-scan accuracy $\eta_p$ and the whole-scan accuracy $\eta_W$ can be given as:
\begin{equation}
\eta_p = \frac{\sum_{s \in S} | R\cap\Gamma_s |}{n_{tot}}, \eta_W = \frac{1}{24} \sum_{s \in S} \frac{| R\cap\Gamma_s |}{n_{\Gamma_s}}
\end{equation}

The total accuracy $\eta_{total}$ is obtained which is comprised of both patch-to-scan and whole-scan accuracies: $\eta_{total} = \eta_p \times \eta_W$.

As for the KIMIA Path960 the accuracy metrics were compliant with leave-one-out approach. Since this is a multi-class dataset,  for each test image, we run it through the the entire training set to obtain the image and its class with the highest probability.  The test and the best-matched image classes are compared to determine if there is a match (i.e. 1) or a mismatch (i.e. 0). The overall accuracy is the percentage of all matched images with respect to the total number of test images (i.e. 960 in this case).

\textbf{Experimentation on Deep Learning --} For KIMIA Path24, we specifically computed four different deep learning structures to compare against the proposed mFLIP descriptor. These deep learning approaches are as follows: (\textit{i}) VGG16: a pre-trained deep net as feature extractor, (\textit{ii}) a fine-tuned VGG16 (transfer learning), (\textit{iii}) Inception V3: a pre-trained deep net as feature extractor, and (\textit{iv}) a fine-tuned Inception V3.

\subsubsection{Pre-Trained CNN as a Feature Extractor}
Specifically for the KIMIA Path24 dataset, the first set of experiments were developed using the Keras library in Python wherein we used pre-trained VGG16 and InceptionV3 for feature extraction without fine-tuning the parameters. In essence, the fully-connected layer (feature vector) for each of these pre-trained models were extracted and provided to an SVM for classification. For linear SVM classification, Python packages scikitlearn and LIBSVM were adopted \cite{pedregosa2011scikit} \cite{chang2011libsvm}. Finally, Python libraries NumPy and SciPy were leveraged to manipulate and store the data \cite{walt2011numpy} \cite{jones2014scipy}.

\subsubsection{Fine-tuned CNN as a Classifier}
For completion, we used the Keras library in Python to fine-tune the pre-trained networks VGG16 and Inception V3 as a classifier against the KIMIA Path24 dataset. For the VGG16 network, we first removed the fully-connected layers from the convolutional layers, after which, we fed the network with training images to extract bottleneck features through the convolutional layers. 
Thereafter, the new fully connected model is attached back onto the VGG16 convolutional layers and trained on each convolutional block, except the last block, in order to receive the adjusted classification weights.

Likewise for the InceptionV3 network, the originally fully connected layer is replaced with a single 1024 dense ReLU layer followed by a softmax classification layer. The new fully connected layers were trained on bottleneck features and then attached back onto the original convolutional layers for training the final two inception blocks. 

\textbf{Evaluation of mFLIP Descriptor --} We performed multiple experiments with different mFLIP configurations in the form of ``mFLIP$_{(L,w,\Delta),D}$'' where $L=|\mathbf{h}|$ is the histogram length, $w$ is the window size, $\Delta$ is the pixel stride (overlap), and $D$ is the distance measure or classification scheme. Specifically, we experimented with $L=127$ and $511$ (after removing the first bin), $w=3$ ($3\times3$), and $\Delta=3$ (no overlap). 

Table \ref{tab:FLIP_measure} provides an overview of the performance of FLIP and mFLIP. When the FLIP is configured with utilizing the original dimensions, with a neighborhood size of 3$\times$3 and $\Delta=3$ pixel stride and a histogram length of $L=127$, the best accuracy ($\eta_{total}$) of $55.24\%$ is achieved using an SVM classifier (with generalized histogram intersection kernel) in the KIMIA Path24 dataset. On the other hand, we obtain a $46.98\%$ accuracy when using histogram intersection distance metric for searching the best-matched image in the KIMIA Path24 dataset -- determined by obtaining the lowest distance when comparing histograms. Currently, the benchmark score for the KIMIA Path24 is achieved by mFLIP - utilizing four image dimensions of $1000\times1000$, $750\times750$, $500\times500$, $250\times250$, with a neighborhood size of 3$\times$3 and $\Delta=3$ pixel stride and a histogram length of $L=508$ (each dimension of which gets a FLIP descriptor of $127$ concatenated together). The best total accuracy in the KIMIA Path24 dataset is an ($\eta_{total}$) of $72.42\%$ which is achieved using an SVM classifier on mFLIP features and a $59.93\%$ accuracy when using $\chi^2$ distance metric (image search). 

\begin{table}[tb]
\centering
\caption{mFLIP and FLIP results for different retrieval strategies ($\chi^2$, histogram intersection, and svm) for a histogram length of $L=127$, generated using neighborhood size of with no-overlap ($\Delta=3$). Best results are highlighted in bold. }
\begin{tabular}{@{}l|ccc}
\hline\hline
& $\eta_{p}$    & $\eta_{W}$    & $\eta_{total}$ \\ \hline
mFLIP$_{(508,3,3),~\chi^2}$   & $77.28$ & $77.55$ & $59.93$    \\
mFLIP$_{(508,3,3),~\textrm{histInt}}$  & $74.87$ & $75.38$ & $56.44$    \\
mFLIP$_{(508,3,3),~\textrm{svm}}$  & $\boldsymbol{84.68}$ & $\boldsymbol{85.52}$ & $\boldsymbol{72.42}$    \\ \hline

FLIP$_{(127,3,3),~\chi^2}$  & $67.62$ & $68.27$ & $46.16$    \\
FLIP$_{(127,3,3),~\textrm{histInt}}$  & $68.07$ & $69.03$ & $46.98$    \\
FLIP$_{(127,3,3),~\textrm{svm}}$  & \textbf{74.11} & \textbf{74.54} & \textbf{55.24}    \\ \hline

\end{tabular}
\label{tab:FLIP_measure}
\end{table} 

After numerous experiments, the best configuration for FLIP is to utilize the highest resolution of the dataset (namely 20x) which results in input images of 1000$\times$1000 equivalent to 0.5$\times$0.5 $mm^2$  that are processed in 3$\times$3 neighbourhood windows with no overlap. Moreover, a 127 bin-size histogram was empirically selected as the size of the FLIP feature vector for each image. A window size of 3$\times$3 is used, as it is the smallest window size that we can utilize for computing the histogram and appears to capture local changes of nuclei and other structures. Additionally, the window of 3$\times$3 was observed to yield the best results when compared against 5$\times$5, 8$\times$8, 32$\times$32, and 64$\times$64 window sizes. Although one has the flexibility to change the window size for any application within the FLIP algorithm, for the purpose of our experimentation with KIMIA Path24, we chose a neighborhood of 3$\times$3 as it yielded the best result.


Table \ref{tab:table_experiment} provides a comparison of FLIP and mFLIP against deep learning methods on the KIMIA Path24 based on gray-scale images. We also show the results of the ELP descriptor that also uses local projections. We explored the performance of a pre-trained deep features versus training from scratch. All the experiments were done on the same KIMIA Path24 dataset. We deduced that pre-trained networks are comparable to training a CNN from scratch. Also, fine-tuning VGG16 does not yield better results despite requiring more training time \cite{kieffer2017convolutional}. We also observed considerable improvement in image search and classification accuracy for the fine-tuned Inception structure. The fine-tuned InceptionV3 delivers $\eta_{total}=56.98$ which is slightly higher than the FLIP accuracy, namely  $\eta_{total}=55.24$. However, all deep learning approaches are considerably lower when compared to the current benchmark, mFLIP$_{(508,3,3)}$ which achieves a $\eta_p=85.53$, $\eta_p=84.68$, and $\eta_{total}=72.42$. The fact that a handcrafted algorithm can surpass deep learning methods, which are the result of substantial design and training efforts, is quite encouraging. However, the reason behind the relatively low performance of  deep features might be due to the feeding of grey scale images to networks while deep networks tend to depend heavily on color. Also the reason for success of mFLIP may be due to the usage of projections in local windows across multiple magnifications. 

For completion, Table \ref{tab:mFLIP_960} provides an overview of the top performing algorithms in the KIMIA Path960 dataset in comparison to the proposed mFLIP algorithm. Although mFLIP does not set the benchmark for the dataset, it certainly competes with the top methods with minimum computation time and resource.


\begin{table}[tb]
\centering
\caption{Results for a SVM classifier on FLIP and mFLIP against the literature.}
\begin{tabular}{lccc} 
\textbf{Method}          & $\eta_{W}$    & $\eta_{p}$    & $\eta_{total}$\\ \hline\hline

mFLIP$_{(508,3,3),\textrm{svm}}$  & $\textbf{85.52}$ & $\textbf{84.68}$ & $\textbf{72.42}$  \\
ELP$_\textrm{svm}$ \cite{elppaper} & $82.70$ & $79.90$ & $66.01$  \\
Inception-v3 (Fine-Tuned) \cite{kieffer2017convolutional} & $76.10$ & $74.87$ &  $56.98$ \\
FLIP$_{(127,3,3),\textrm{svm}}$  & \textbf{74.54} & \textbf{74.11} & \textbf{55.24}  \\ 
Inception-v3 (Feature Extractor) \cite{kieffer2017convolutional} & $71.24$ & $70.94$ &  $50.54$ \\
VGG+RF \cite{bizzego2019evaluating}   & $67.12$    & $64.66$    & $43.40$  \\ 
VGG16 (Feature Extractor) \cite{kieffer2017convolutional} & $64.96$ & $65.21$ &  $42.36$ \\
VGG16 (Fine-Tuned) \cite{kieffer2017convolutional} & $66.23$ & $63.85$ &  $42.29$ \\
CNN (Trained from Scratch) \cite{babaie2017classification}   & $64.75$    & $64.98$    & $41.80$  \\ 

\hline
		
		\hline\hline
	\end{tabular}
	\label{tab:table_experiment}
\end{table}

\begin{table}[htb]
\centering
\caption{mFLIP accuracy against other methods for KimiaPath960.}
\begin{tabular}{lc}
\hline\hline
\textbf{Method}             & \textbf{Accuracy} \\ \hline
BoVW$_\textrm{(1200 codebooks),~\textrm{IKSVM}}$ \cite{kumar2017comparative}\qquad & $94.87$    \\
VGG16$_{~\textrm{L2}}$ \cite{kumar2017comparative}        & $94.72$    \\
AlexNet$_{~\textrm{L1}}$ \cite{kumar2017comparative}      & $91.35$    \\
LBP$_{~\textrm{L2}}$ \cite{kumar2017comparative}          & $90.62$    \\
mFLIP$_{~\chi^2}$       & $88$     \\
mFLIP$_{~\textrm{svm}}$       & $87$    \\ 
\hline \hline
\end{tabular}
\label{tab:mFLIP_960}
\end{table} 

\section{Conclusions}
\label{sec:conclusion}
Here we introduced a new feature descriptor called \emph{Forming Local Intersections of Projections} (FLIP) wherein we have shown that using element-wise intersections of local Radon projections, followed by re-scaling to create a histogram, can be used to construct a new image descriptor. In addition, a multi-resolution FLIP descriptor (mFLIP) is also introduced and validated against the publicly available, KIMIA Path24 and KIMIA Path960 datasets. Specifically, the mFLIP is observed to outperform deep solutions when tested on the KIMIA Path24 dataset. Furthermore, both FLIP and mFLIP provide a more compact image representation with 128 and 508 bins, respectively, compared to generally high-dimensionality of deep features (i.e., 4096 for CNN and VGG16) in the KIMIA Path24 dataset. It appears that mFLIP is particularly suitable for histopathology images as the proposed algorithm is observed to capture the texture of each image through the means of Radon projections and to quantify these projections onto a condensed histogram. In addition, the process of localizing and capturing the Radon transform for small neighborhood does not require learning or expensive training. The novel image descriptor (FLIP), and its multi resolutional version,the mFLIP descriptor have surpassed the current benchmark for the KIMIA Path24 dataset by achieving a total accuracy of $\approx 72\%$ using an SVM classification with generalized histogram intersection kernel. We must mention that we have processed gray-scale images. Therefore, crucial information, such as staining that has chemical meaning in histopathology, may have been lost. 

\bibliography{references}
\end{document}